\documentclass[runningheads,a4paper]{llncs}

\usepackage{graphicx}
\usepackage[sectionbib]{natbib}
\usepackage{multirow}
\usepackage{amsmath}         
\usepackage{amssymb}  
\usepackage{url}

\newcommand{\keywords}[1]{\par\addvspace\baselineskip
\noindent\keywordname\enspace\ignorespaces#1}

\long\def\symbolfootnote[#1]#2{
\begingroup
	\def\thefootnote{\fnsymbol{footnote}}\footnote[#1]{#2}
\endgroup}

\begin{document}

\mainmatter  

\title{The Relation Between Acausality and Interference in Quantum-Like Bayesian Networks}

\titlerunning{Lecture Notes in Computer Science}

\author{Catarina Moreira \and Andreas Wichert}
				   	
\institute{Instituto Superior T\'{e}cnico, INESC-ID\\Av. Professor Cavaco Silva, 2744-016 Porto Salvo, Portugal\\
 \email{$\{$catarina.p.moreira, andreas.wichert$\}$@ist.utl.pt  }
}

\maketitle

\begin{abstract}

We analyse a quantum-like Bayesian Network that puts together cause/effect relationships and semantic similarities between events. These semantic similarities constitute acausal connections according to the Synchronicity principle and provide new relationships to quantum like probabilistic graphical models. As a consequence, beliefs (or any other event) can be represented in vector spaces, in which quantum parameters are determined by the similarities that these vectors share between them. Events attached by a semantic meaning do not need to have an explanation in terms of cause and effect. 

\keywords{Quantum Cognition; Quantum-Like Bayesian Networks; Synchronicity Principle}
\end{abstract}

\symbolfootnote[0]{This work was supported by national funds through Funda\c{c}\~{a}o para a Ci\^{e}ncia e Tecnologia (FCT) with reference UID/CEC/50021/2013 and through the PhD grant SFRH/BD/92391/2013.}

\section{Introduction}

Current decision-making systems face high levels of uncertainty resulting from data, which is either missing or untrustworthy. These systems usually turn to probability theory as a mathematical framework to deal with uncertainty. One problem, however, is that it is hard  for these systems to make reliable predictions in situations where the laws of probability are being violated. These situations happen quite frequently in systems which try to model human decisions~\citep{Tversky74,Tversky83Uncertainty,Tversky92}. 

Uncertainty in decision problems arises, because of limitations in our ability to observe the world and in limitations in our ability to model it~\citep{koller09prob}. If we could have access to all observations of the world and extract all the information it contained, then one could have access to the full joint probability distribution describing the relation between every possible random variable. This knowledge would eliminate uncertainty and would enable any prediction. This information, however, is not available and not possible to obtain as a full, leading to uncertainty. A formal framework capable of representing multiple outcomes and their likelihoods under uncertainty is probability theory~\citep{Murphy12}.

In an attempt to explain the decisions that people make under risk, cognitive scientists started to search for other mathematical frameworks that could also deal with uncertainty. Recent literature suggests that quantum probability can accommodate these violations and improve the probabilistic inferences of such systems~\citep{Aerts95quantum_structures,busemeyer06,Bordley98}. 

Quantum cognition is a research field that aims at using the mathematical principles of quantum mechanics to model cognitive systems for human decision making~\citep{Busemeyer15,Busemeyer14,Aerts14}. Given that Bayesian probability theory is very rigid in the sense that it poses many constraints and assumptions (single trajectory principle, obeys set theory, etc.), it becomes too limited to provide simple models that can capture human judgments and decisions, since people are constantly violating the laws of logic and probability theory~\citep{Tversky74,Tversky83Uncertainty,Tversky92}. Recent literature suggests that quantum probability can be used as a mathematical alternative to the classical theory and can accommodate these violations~\citep{Mura09,Mogiliansky09,Aerts11}. It has been showed that quantum models provide significant advantages towards classical models~\citep{Busemeyer15decision,Busemeyer12hierarchical}.

In this work, we explore the implications of causal relationships in quantum-like probabilistic graphical models and also the implications of semantic similarities between quantum events~\citep{Moreira15}. These semantic similarities provide new relationships to the graphical models and enables the computation of quantum parameters through vector similarities.

This work is organised as follows. In Sections 2 and 3, we address two types of relationships, respectively: cause/effect and acausal relationships. In Section 4, we describe a quantum-like Bayesian Network that takes advantages of both cause/effect relationships and semantic similarities (acausal events). In Section 5, we show and analyse the applications of the proposed model in current decision problems. Finally, in Section 6, we conclude with some final remarks regarding the application of quantum-like Bayesian Networks to decision problems.

\section{What is Causation}

Most events are reduced to the principle of causality, which is the connection of phenomena where the cause gives rise to some effect. This is the philosophical principle that underlies our conception of natural law~\citep{Jung12}.  

Under the principle of causality, some event $A$ can have more than one cause, in which none of them alone is sufficient to produce $A$. Causality is usually: (1) transitive, if some event $A$ is a cause of $B$  and $B$ is a cause of $C$, then $A$ is also a cause of $C$; (2)  irreflexible, an event $A$ cannot cause itself; and (3) antisymmetric, if $A$ is a cause of $B$, then $B$ is not a cause of $A$~\citep{spirtes00}.

The essence of causality is the generation and determination of one phenomenon by another. Causality enables the representation of our knowledge regarding a given context through {\it experience}. By experience, we mean that  the observation of the relationships between events enables the detection of irrelevancies in the domain. This will lead to the construction of causal models with minimised relationships between events~\citep{Pearl88}. Bayesian Networks are examples of such models.

Under the principle of causality, two events that are not causally connected should not produce any effects. When some acausal events occur by producing an effect, it is called a coincidence. Carl Jung, believed that nothing happens by chance and, consequently, all events had to be connected between each other, not in a causal setting, but rather in a meaningful way. Under this point of view, Jung proposed the Synchronicity principle~\citep{Jung12}.  

\section{Acausal Connectionist Principle}

The Synchronicity principle may occur as a single event of a chain of related events and can be defined by a significant coincidence which appears between a mental state and an event occurring in the external world~\citep{Martin09}. Jung believed that two acausal events did not occur by chance, but rather by a shared meaning. Therefore, in order to experience a synchronised event, one needs to extract the meaning of its symbols for the interpretation of the synchronicity.  So, the Synchronicity principle can be seen as a correlation between two acausal events which are connected through meaning~\citep{Jung12}.   

Jung defended that the connection between a mental state and matter is due to the energy emerged from the emotional state associated to the synchronicity event~\citep{Jung12}. This metaphysical assertion was based on the fact that it is the person's interpretation that defines the meaning of a synchronous event. This implies a strong relation between the extraction of the semantic meaning of events and how one interprets it. If there is no semantic extraction, then there is no meaningful interpretation of the event, and consequently, there is no synchronicity~\citep{Lindorff04}.

It is important to mention that the Synchronicity principle is a concept that does not question or compete with the notion of causality. Instead, it maintains that just as events may be connected by a causal line, they may also be connected by meaning. A grouping of events attached by meaning do not need to have an explanation in terms of cause and effect. 

In this work, we explore the consequences of the synchronicity principle applied to quantum states with high levels of uncertainty as a way to provide additional information to quantum-like probabilistic graphical models, which mainly contain cause/effect relationships. Although the principles of probability are well established, such that synchronicity might be seen as the occurrence of coincidences, in the quantum mechanics realm, given the high levels of uncertainty that describe the quantum states, the coincidences or improbable occurrences happen quite often.

\section{Quantum-Like Bayesian Networks: Combining Causal and Acausal Principles for Quantum Cognition}

The reason why we are turning to Bayesian Networks is because they are inspired in human cognition~\citep{Griffiths08}. It is easier for a person to combine pieces of evidence and to reason about them, instead of calculating all possible events and their respective beliefs. In the same way, Bayesian Networks also provide this link between human cognition and rational inductive inference. Instead of representing the full joint distribution, Bayesian Networks represent the decision problem in small modules that can be combined to perform inferences. Only the probabilities which are actually needed to perform the inferences are computed.

\subsection{Classical Bayesian Networks}\label{sec:cbn}

A classical Bayesian Network is a directed acyclic graph structure. Each node represents a different random variable from a specific domain and each edge represents a direct influence from the source node to the target node. The graph also represents independence relationships between random variables and is followed by a conditional probability table which specifies the probability distribution of the current node given its parents~\citep{koller09prob}.

A Bayesian Network represents a full joint probability distribution through conditional independence statements in order to answer queries about the domain. The full joint distribution~\citep{russel10} of a Bayesian Network, where $X$ is the list of variables, is given by Equation~\ref{eq:joint}.
\begin{equation}
Pr_c( X_1, \dots, X_n ) = \prod_{i=1}^n  Pr( X_i | Parents(X_i) ) 
\label{eq:joint}
\end{equation}
In order to answer queries, the network enables the combination of the relevant entries of the full joint probability distribution. This process consists in the computation of the marginal probability distribution of the network. Let $e$ be the list of observed variables and let $Y$ be the remaining unobserved variables in the network. For some query $X$, the inference is given by Equation~\ref{eq:inference}.
\begin{equation}
Pr_c( X | e ) = \alpha \left[ \sum_{y \in Y} Pr_c( X, e, y) \right]  \text{Where~~~}\alpha = \frac{1}{ \sum_{x \in X} Pr_c(X = x, e) }
\label{eq:inference}
\end{equation} 
The summation is over all possible $y$, i.e., all possible combinations of values of the unobserved variables $y$. The $\alpha$ parameter, corresponds to the normalisation factor for the distribution $Pr(X | e)$~\citep{russel10}. This normalisation factor comes from some assumptions that are made in Bayes rule.

\subsection{From Classical Bayesian Networks to Quantum-Like Networks}

Suppose that we have a Bayesian Network with three random variables with the following structure: $B \leftarrow A \rightarrow C $. In order to determine the probability of node B, we would need to make the following computation based on Equation~\ref{eq:inference}.
\[ Pr( B = t ) = Pr( A = t ) Pr( B = t | A = t ) Pr( C = t | A = t ) + \] 
\[ ~~~~~~~~~~~~~ + Pr( A = t ) Pr( B = t | A = t ) Pr( C = f | A = t ) + \] 
\[ ~~~~~~~~~~~~~ + Pr( A = f ) Pr( B = t | A = f ) Pr( C = t | A = f ) + \]  
\begin{equation}
~~~~~~~~~~~~~ + Pr( A = f ) Pr( B = t | A = f ) Pr( C = f | A = f ) 
\label{eq:bn_1}
\end{equation}
A classical probability can be converted into a quantum probability amplitude in the following way. Suppose that events $A_1, \dots, A_N$ form a set of mutually disjoint events, such that their union is all in the sample space, $\Omega$, for any other event $B$.  The classical law of total probability can be formulated like in Equation~\ref{eq:law_total_prob_c}.
\begin{equation}
Pr(B) = \sum_{i=1}^{N} Pr(A_i) Pr(B | A_i)   \text{~~~~~~~~~~where:~} \sum_{i = 1}^{N} Pr(A_i) = 1
\label{eq:law_total_prob_c}
\end{equation}
The quantum law of total probability can be derived through Equation~\ref{eq:law_total_prob_c} by applying Born's rule~\citep{Caves02,Nielsen00}:
\begin{equation}
Pr(B) = \left| \sum_{j=1}^N e^{i\theta_j}\psi_{A_j} \psi_{B | A_j} \right|^2 \text{~~~~~~~~where:~} \sum_{j= 1}^{N} \left| e^{i\theta_j} \psi_{A_j} \right|^2  = 1
\label{eq:law_total_prob_q}
\end{equation}
Returning to our example, in order to convert the real probabilities in Equation~\ref{eq:bn_1} into quantum amplitudes, one needs to apply Born's rule. In Equation~\ref{eq:qbn_1}, the term $ \psi_{1} e^{\theta_1}$ corresponds to the quantum probability amplitude of the term $Pr( A = t ) Pr( B = t | A = t ) Pr( C = t | A = t )$; the term $ \psi_{2} e^{\theta_2}$ corresponds to the quantum probability amplitude of the term $Pr( A = t ) Pr( B = t | A = t ) Pr( C = f | A = t )$ and so on.
\[ Pr( B = t ) = \left| \psi_{1} e^{\theta_1} + \psi_{2} e^{\theta_2}  + \psi_{3} e^{\theta_3}   + \psi_{4} e^{\theta_4} \right|^2  \] 
\begin{equation}
\label{eq:qbn_1}
\end{equation}
Expanding Equation~\ref{eq:qbn_1},
\[ Pr( B = t ) = \left|  \psi_{1} e^{\theta_1} \right|^2 + \left|  \psi_{2} e^{\theta_2} \right|^2  + \left|  \psi_{3} e^{\theta_3} \right|^2 + \left|  \psi_{4} e^{\theta_4} \right|^2 +  \left|  \psi_{1} e^{\theta_1} \right| \left|  \psi_{2} e^{\theta_2} \right| +  \left|  \psi_{2} e^{\theta_2} \right| \left|  \psi_{1} e^{\theta_1} \right| +  \] 
\[  +  \left|  \psi_{1} e^{\theta_1} \right| \left|  \psi_{3} e^{\theta_3} \right| +  \left|  \psi_{3} e^{\theta_3} \right| \left|  \psi_{1} e^{\theta_1} \right| +  \left|  \psi_{1} e^{\theta_1} \right| \left|  \psi_{4} e^{\theta_4} \right| + \left|  \psi_{4} e^{\theta_4} \right| \left|  \psi_{1} e^{\theta_1} \right| + \] 
\[ +  \left|  \psi_{2} e^{\theta_2} \right| \left|  \psi_{3} e^{\theta_3} \right| + \left|  \psi_{3} e^{\theta_3} \right| \left|  \psi_{2} e^{\theta_2} \right| +  \left|  \psi_{2} e^{\theta_2} \right| \left|  \psi_{4} e^{\theta_4} \right| + \left|  \psi_{4} e^{\theta_4} \right| \left|  \psi_{2} e^{\theta_2} \right| + \] 
\begin{equation}
+  \left|  \psi_{3} e^{\theta_3} \right| \left|  \psi_{4} e^{\theta_4} \right| + \left|  \psi_{4} e^{\theta_4} \right| \left|  \psi_{3} e^{\theta_3} \right|~~~~~~~~~~~~~~~~~~~~~~~~~~~~~~~~~~~~~~~~~~~~~
\label{eq:qbn_2}
\end{equation}
Knowing that, $2 \cos( \theta_1 - \theta_2) = e^{i\theta_1- i\theta_2} + e^{i\theta_2- i\theta_1}$, then Equation~\ref{eq:qbn_2} becomes:
\[ Pr( B = t ) = \sum_i^N \left|  \psi_{i}  \right|^2  + 2 \left|  \psi_{1} \right| \left|  \psi_{2} \right| \cos( \theta_1 - \theta_2 ) + 2 \left|  \psi_{1} \right| \left|  \psi_{3} \right| \cos( \theta_1 - \theta_3 ) +  \] 
\begin{equation}  
+ 2 \left|  \psi_{1} \right| \left|  \psi_{4} \right| \cos( \theta_1 - \theta_4 ) + 2 \left|  \psi_{2} \right| \left|  \psi_{3} \right| \cos( \theta_2 - \theta_3 )  + \dots + 2 \left|  \psi_{3} \right| \left|  \psi_{4} \right| \cos( \theta_3 - \theta_4 )
\label{eq:qbn_2}
\end{equation}
Equation~\ref{eq:qbn_2} can be rewritten as:
\begin{equation}  
Pr( B = t ) = \sum_i^N \left|  \psi_{i}  \right|^2  + 2 \sum_{i=1}^{N-1} \sum_{j=i+1}^{N} \left|  \psi_{i} \right| \left|  \psi_{j} \right| \cos( \theta_i - \theta_j ) 
\label{eq:qbn_3}
\end{equation}

\subsection{Quantum-Like Bayesian Network}\label{sec:qlbn}

A quantum-like Bayesian Network can be defined in the same way as a classical Bayesian Network with the difference that real probability numbers are replaced by quantum probability amplitudes~\citep{Tucci95,Leifer08}. 

The quantum counterpart of the full joint probability distribution corresponds to the application of Born's rule to Equation~\ref{eq:joint}. This results in Equation~\ref{eq:joint_q}, where $QPr$ corresponds to a quantum amplitude.
\begin{equation}
Pr_q( X_1, \dots, X_n ) = \left| \prod_{i = 1}^n QPr( X_i | Parents( X_i ) ) \right|^2
\label{eq:joint_q}
\end{equation}
When performing probabilistic inferences in Bayesian Networks, the probability amplitude of each assignment of the network is propagated and influences the probabilities of the remaining nodes. In order to perform inferences on the network, one needs to apply Born's rule to the classical marginal probability distribution, just like in was presented in Equation~\ref{eq:qbn_3}. If we rewrite this equation with the notation presented in Equation~\ref{eq:bn_1}, then the quantum counterpart of the classical marginalization formula for inferences in Bayesian Networks becomes:
\begin{equation}
Pr_{q}(X | e) = \alpha \sum_{i = 1}^{|Y|}  \left| \prod_x^N  QPr( X_x | Parents(X_x), e, y = i )  \right| ^2 + 2 \cdot Interference
\label{eq:final1}
\end{equation}
\begin{multline*}
Interference =\\  \sum_{i=1}^{|Y|-1} \sum_{j=i+1}^{|Y|}  \left| \prod_x^N QPr( X_x | Parents(X_x), e, y=i ) \right| \cdot \\  \left| \prod_x^N  QPr( X_x | Parents(X_x), e, y= j ) \right| \cdot \cos( \theta_i - \theta_j )
\end{multline*}
In classical Bayesian inference, normalisation of the inference scores is necessary due to the independence assumptions made in Bayes rule. In quantum-like inferences, we need to normalize the final scores, not only because of the asme independence assumptions, but also because of the quantum interference term. If the conditional probability tables of the proposed quantum-like Bayesian Network were double stochastic, then this normalization would not be necessary. But, since in the proposed model we do not have this constraint, then a normalization is required after the computation of the probabilistic inference.

Following Equation~\ref{eq:final1}, when $ \cos( \theta_i - \theta_j) $  equals zero, then it is straightforward that quantum probability theory converges to its classical counterpart, because the interference term will be zero. 

For non-zero values, Equation~\ref{eq:final1} will produce interference effects that can affect destructively the classical probability (when the interference term in smaller than zero) or  constructively (when it is bigger than zero). Additionally,  Equation~\ref{eq:final1} will lead to a large amount of $\theta$ parameters when the number of events increases. For $N$ binary random variables, we will end up with $2^ {N} $ parameters.

\subsection{Semantic Networks: Incorporating Acausal Connections}

A semantic network is often used for knowledge representation. It corresponds to a directed or undirected graph in which nodes represent concepts and edges reflect semantic relations. The extraction of the semantic network from the original Bayesian Network is a necessary step in order to find variables that are only connected in a meaningful way (and not necessarily connected by cause/effect relationships), just like it is stated in the Synchronicity principle.

\begin{figure}[ht]
	\centering
	\includegraphics[scale=0.3]{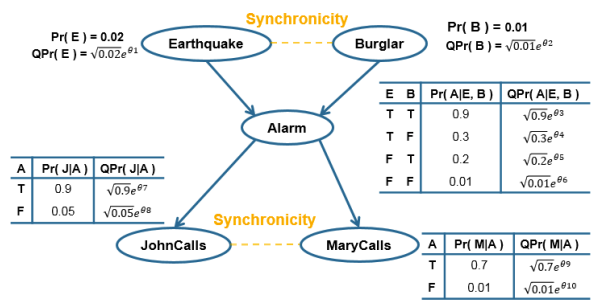}
	\caption{Example of a Quantum-Like Bayesian Network~\citep{russel10}. $QPr$ represents quantum  amplitudes. $Pr$ corresponds to the real classical probabilities.}
	\label{fig:structure}
\end{figure}

\begin{figure}[ht]
	\centering
	\includegraphics[scale=0.3]{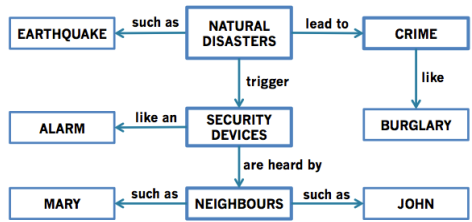}
	\caption{Semantic Network representation of the network in Figure~\ref{fig:structure}}
	\label{fig:semantic_web}	
\end{figure}

Consider the Bayesian Network in Figure~\ref{fig:structure}~\citep{russel10,Pearl88}. In order to extract its semantic meaning, we need to take into account the context of the network. Suppose that you have a new burglar alarm installed at home. It can detect burglary, but also sometimes responds to earthquakes. John and Mary are two neighbours, who promised to call you when they hear the alarm. John always calls when he hears the alarm, but sometimes confuses telephone ringing with the alarm and calls too. Mary likes loud music and sometimes misses the alarm. 
From this description, we extracted the semantic network, illustrated in Figure~\ref{fig:semantic_web}, which represents the meaningful connections between concepts. The following knowledge was extracted. It is well known that catastrophes cause panic among people and, consequently, increase crime rates, more specifically burglaries. So, a new pair of synchronised variables between \emph{Earthquake} and \emph{Burglar} emerges. Moreover, \emph{John} and \emph{Mary} derive both from the same concept $person$, so, these two nodes will also be synchronised. These synchronised variables mean that, although there is no explicit causal connection between these nodes in the Bayesian Network, they can become correlated through their meaning. 
\subsection{Setting Quantum Parameters According to the Synchronicity Principle}

\begin{figure}[h!]
\resizebox{\columnwidth}{!} {
\centering
\includegraphics[scale = 0.3]{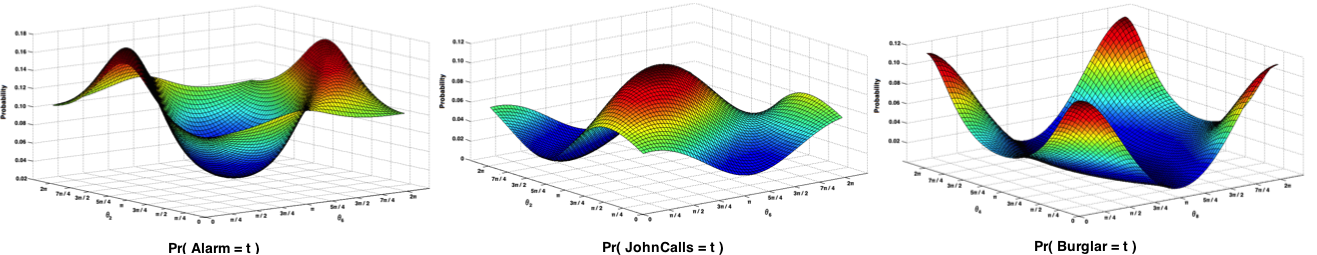}
}
\caption{Variation of the probability values of the Bayesian Network in Figure~\ref{fig:structure} for different quantum parameters~\citep{Moreira14}.}
\label{fig:interval}
\end{figure}

In Section~\ref{sec:qlbn}, it was presented that Equation~\ref{eq:final1} generates an exponential number of quantum $\theta$ parameters according to the number of unknown variables. If nothing is told about how to assign these quantum parameters, then we end up with an interval of possible probabilities. For instance, Figure~\ref{fig:interval} shows that, the probabilities for the different random variables of the Quantum-Like bayesian Network from~\cite{Moreira14} can range from an interval of possible probability values. This means that one needs some kind of heuristic function that is able to assign these quantum parameters automatically. 
We define the Synchronicity heuristic in a similar way to Jung's principle: two variables are said to be synchronised, if they share a meaningful connection between them. This meaningful connection can be obtained through a semantic network representation of the variables in question. This will enable the emergence of new meaningful connections that would be inexistent when considering only cause/effect relationships. The quantum parameters are then tuned in such a way that the angle formed by these two variables, in a Hilbert space, is the smallest possible, this way forcing acausal events to be correlated. 

\begin{figure}
\centering
\includegraphics[scale=0.6]{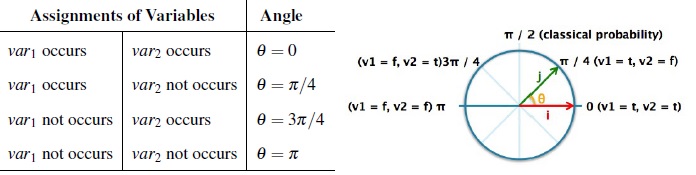}
\caption{Encoding of the Synchronized variables with their respective angles (left). Two synchronized events forming an angle of $\pi/4$ between them (right).}
\label{fig:synchr}
\end{figure}

For the case of binary variables, the Synchronicity heuristic is associated with a set of two variables, which can be in one of four possible states. The Hilbert space is partitioned according to these four states, as exemplified in Figure~\ref{fig:synchr}. The angles formed by the combination of these four possible states are detailed in the table also in Figure~\ref{fig:synchr}~\citep{Moreira15}.
In the right extreme of the Hilbert space represented in Figure~\ref{fig:synchr}, we encoded it as the occurrence of a pair of synchronised variables. So, when two synchronised variables occur, the smallest angle that these vectors make between each other corresponds to $\theta = 0$. The most dissimilar vector corresponds to the situation where two synchronised variables do not occur. So, we set $\theta$ to be the largest angle possible, that is $\pi$. 

The other situations correspond to the scenarios where one synchronised variable occurs and the other one does not. In Figure~\ref{fig:synchr}, the parameter $\theta$ is chosen according to the smallest angle that these two vectors, $i$ and $j$, make between each other, that is $\pi/4$. We are choosing the smallest angle, because we want to correlate these two acausal events by forcing the occurrence of \emph{coincidences} between them, just like described in the Synchronicity principle. The axis corresponding to $\pi/2$ and $3\pi/2$ were ignored, because they correspond to classical probabilities ($\cos \left(  \pi/2 \right) = \cos \left(3\pi/2 \right) = 0 $). We are taking steps of $\pi/4$ inspired by the quantum law of interference proposed by~\cite{Yukalov11}, in which the authors suggest to replace the quantum interference term by $1/4$.

\section{Example of Application}

\begin{figure}[ht]
\centering
\includegraphics[scale = 0.4]{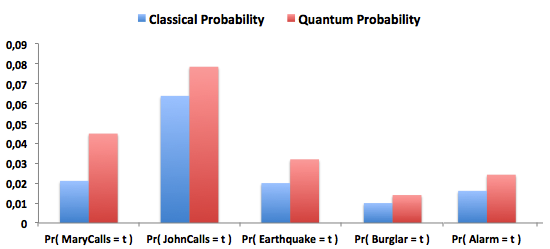}
\caption{Results for various queries comparing probabilistic inferences using classical and quantum probability when no evidences are observed: maximum uncertainty.}
\label{fig:results1}
\end{figure}

We queried each variable of the network in Figure~\ref{fig:structure} without providing any observation. We performed the following queries: $Pr(JonhCalls = true)$, $Pr(MaryCalls = true)$, $Pr(Alarm = true)$, $Pr(Burglar = true)$ and $Pr(Earthquake = true)$. We then extracted both classical and quantum inferences and represented the results in the graph in Figure~\ref{fig:results1}.

Figure~\ref{fig:results1}, shows that, when nothing is known about the state of the world, quantum probabilities tend to increase and overcome their classical counterpart. In quantum theory, when nothing is observed, all nodes of the Bayesian Network are in a superposition state. For each possible configuration in this superposition state, a probability amplitude is associated to it. During the superposition state, the amplitudes of the probabilities of the nodes of the Bayesian Network start to be modified due to the interference effects. If one looks at the nodes as waves crossing the network from different locations, these waves can crash between each other, causing them to be either destroyed or to be merged together. This interference of the waves is controlled through the Synchronicity principle by linking acausal events.

When one starts to provide information to the Bayesian Network, then the superposition state collapses into another quantum state, affecting the configuration of the remaining possible states of the network. Moreover, by making some observation to the network, we are reducing the total amount of uncertainty and, consequently, the reduction of the waves crossing the network (Table~\ref{tab:results_one_ev}). 
\begin{table}[ht]
\resizebox{\columnwidth}{!} {
\begin{tabular}{l | l | c | c | c | c | c | c}
	&\bf{Evidences}	& \bf{Pr( Alarm = t )} 		& \bf{Pr( Earthquake = t )} & \bf{Pr( Burglar = t )}	& \bf{Pr( JohnCalls = t )}	& \bf{Pr( MaryCalls = t )} \\
	\hline		

 \parbox[c]{2mm}{\multirow{5}{*}{\rotatebox[origin=c]{90}{\bf{CLASSIC}}}} 
&\bf{JohnCalls = t}		& 0.2277 		& 0.0949 		& 0.1333		& 1.0000 		& 0.1671	\\
&\bf{MaryCalls = t}		& 0.5341 		& 0.2033 		& 0.3119		& 0.5040 		& 1.0000	\\
&\bf{Earthquake = t}	& 0.2966		& 1.0000 		& 0.0100		& 0.3021 		& 0.2147	\\	
&\bf{Burglar = t}		& 0.9402		& 0.0200 		& 1.0000		& 0.8492 		& 0.6587	\\		
&\bf{Alarm = t}			& 1.0000		& 0.3581 		& 0.5835		& 0.9000 		& 0.7000	\\
\hline	
\hline
 \parbox[c]{2mm}{\multirow{5}{*}{\rotatebox[origin=c]{90}{~\bf{QUANTUM}}}}
&\bf{JohnCalls = t}		& 0.3669		& 0.1484 		& 0.2124 		& 1.0000 		& 0.2321	 \\	
&\bf{MaryCalls = t}		& 0.6598 		& 0.2239		& 0.3474 		& 0.6032 		& 1.0000 \\
&\bf{Earthquake = t}	& 0.4389		& 1.0000 		& 0.0124 		& 0.4012 		& 0.2403	 \\	
&\bf{Burglar = t}		& 0.9611 		& 0.02	 	& 1.0000 		& 0.8583 		& 0.6337	 \\	
&\bf{Alarm = t}			& 1.0000		& 0.3431 		& 0.5560		& 0.9000 		& 0.7000 \\
\hline	
\end{tabular}
}
\caption{Probabilities obtained when performing inference on the Bayesian Network of Figure~\ref{fig:structure}. }
\label{tab:results_one_ev}
\end{table}

In Table~\ref{tab:results_one_ev} there are two pairs of synchronised variables: (Earthquake, Burglar) and (MaryCalls, JohnCalls). The quantum probability of $Pr( Earthquake = t | JohnCalls = t)$ has increased almost the same quantity as for the probability $Pr( Burglar = t | JohnCalls = t)$ (56.37\% for earthquake and 59.34\% for burglar). In the same way, when we observe that $MaryCalls = t$, then the percentage of a Burglary increased 11.38\%, whereas Earthquake increased a percentage of 10.13\% towards its classical counterpart. 

\section{Conclusions}

In this work, we analysed a quantum-like Bayesian Network that puts together cause/effect relationships and semantic similarities between events. These similarities constitute acausal connections according to the Synchronicity principle and provide new relationships to the graphical models. As a consequence,  events can be represented in vector spaces, in which quantum parameters are determined by the similarities that these vectors share between them.  In the realm of quantum cognition, quantum parameters might represent the correlation between events (beliefs) in a meaningful acausal relationship. 

The proposed quantum-like Bayesian Network benefits from the same advantages of classical Bayesian Networks: (1) it enables a visual representation of all relationships between all random variables of a given decision scenario, (2) can perform inferences over unobserved variables, that is, can deal with uncertainty, (3) enables the detection of independent and dependent variables more easily. Moreover, the mapping to a quantum-like approach leads to a new mathematical formalism for computing inferences in Bayesian Networks that takes into account quantum interference effects. These effects can accommodate puzzling phenomena that could not be explained through a classical Bayesian Network. This is probably the biggest advantage of the proposed model. A network structure that can combine different sources of knowledge in order to model a more complex decision scenario and accommodate violations to the Sure Thing Principle. 

With this work, we argue that, when presented with a problem, we perform a semantic categorisation of the symbols that we extract from the given problem through our thoughts~\cite{Osherson95}. Since our thoughts are abstract, cause/effect relationships might not be the most appropriate mechanisms to simulate interferences between them. The Synchronicity principle seems to fit more in this context, since our thoughts can relate to each other from meaningful connections, rather than cause/effect relationships~\citep{Jung12}.

We end this work with some reflections. Over the literature of quantum cognition, quantum models have been proposed in order to explain some paradoxical findings~\citep{Busemeyer09,Haven13}. These decision problems, however, are very small. They are modelled with at most two random variables. Decision problems with more random variables suffer from the problem of the exponential generation of quantum parameters (like in Burglar/Alarm Bayesian Network). For more complex problems, how can one model them, since the only apparent way to do so, is through the usage of heuristic functions that can assign values to the quantum $\theta$ parameters? But even through this method, given the lack of experimental data, how can one validate such functions? Is the usage of these functions a correct way to tackle this problem, or is it wrong to proceed in this direction? How can such experiment be conducted? Is it even possible to show violations on the laws of probability theory for more complex problem?  

\bibliographystyle{apalike}

\end{document}